\pdfoutput=1

\PassOptionsToPackage{dvipsnames}{xcolor}
\documentclass[11pt]{article}

\usepackage{etoolbox}
\newcommand{\zerodisplayskips}{%
  \setlength{\abovedisplayskip}{0pt}%
  \setlength{\belowdisplayskip}{2pt}%
  \setlength{\abovedisplayshortskip}{0pt}%
  \setlength{\belowdisplayshortskip}{0pt}}
\appto{\normalsize}{\zerodisplayskips}
\appto{\small}{\zerodisplayskips}
\appto{\footnotesize}{\zerodisplayskips}

\usepackage{authblk}

\usepackage[]{emnlp2021}

\usepackage{times}
\usepackage{latexsym}
\usepackage{multirow}
\usepackage{graphicx}

\usepackage{mwe}
\usepackage{amsmath}

\usepackage[T1]{fontenc}

\usepackage[utf8]{inputenc}

\usepackage{microtype}

\title{Learning to Perform Complex Tasks through\\Compositional Fine-Tuning of Language Models}

\author[1]{{\bf Victor S. Bursztyn}}
\author[1]{{\bf David Demeter}}
\author[1,2]{{\bf Doug Downey}}
\author[1]{{\bf Larry Birnbaum}}
\affil[1]{Department of Computer Science, Northwestern University, Evanston, IL, USA}
\affil[2]{Allen Institute for Artificial Intelligence, Seattle, WA, USA}
\affil[ ]{\texttt{\{v-bursztyn,ddemeter\}@u.northwestern.edu}}
\affil[ ]{\texttt{\{d-downey,l-birnbaum\}@northwestern.edu}}

\begin{document}
\maketitle
\begin{abstract}
How to usefully encode compositional task structure has long been a core challenge in AI. Recent work in chain of thought prompting has shown that for very large neural language models (LMs), explicitly demonstrating the inferential steps involved in a target task may improve performance over end-to-end learning that focuses on the target task alone. However, chain of thought prompting has significant limitations due to its dependency on huge pretrained LMs. In this work, we present compositional fine-tuning (CFT): an approach based on explicitly decomposing a target task into component tasks, and then fine-tuning smaller LMs on a curriculum of such component tasks. We apply CFT to recommendation tasks in two domains, world travel and local dining, as well as a previously studied inferential task (sports understanding). We show that CFT outperforms end-to-end learning even with equal amounts of data, and gets consistently better as more component tasks are modeled via fine-tuning. Compared with chain of thought prompting, CFT performs at least as well using LMs only 7.4\% of the size, and is moreover applicable to task domains for which data are not available during pretraining.
\end{abstract}

\section{Introduction}
\label{sec:intro}

Philosophy, linguistics, and computer science have long debated how and whether to explicitly encode the compositionality of task structure in models of language understanding and generation \cite{fodor1988connectionism}. The prevailing paradigm in today's NLP is {\em end-to-end learning}, in which the learning of compositional task structure is subsumed by the learning of a complex target task, with the support of increasingly powerful language models (LMs) \cite{devlin2019bert, raffel2020exploring, brown2020language}.

\begin{figure}[htb]
  \centering
  \includegraphics[width=.45\textwidth]{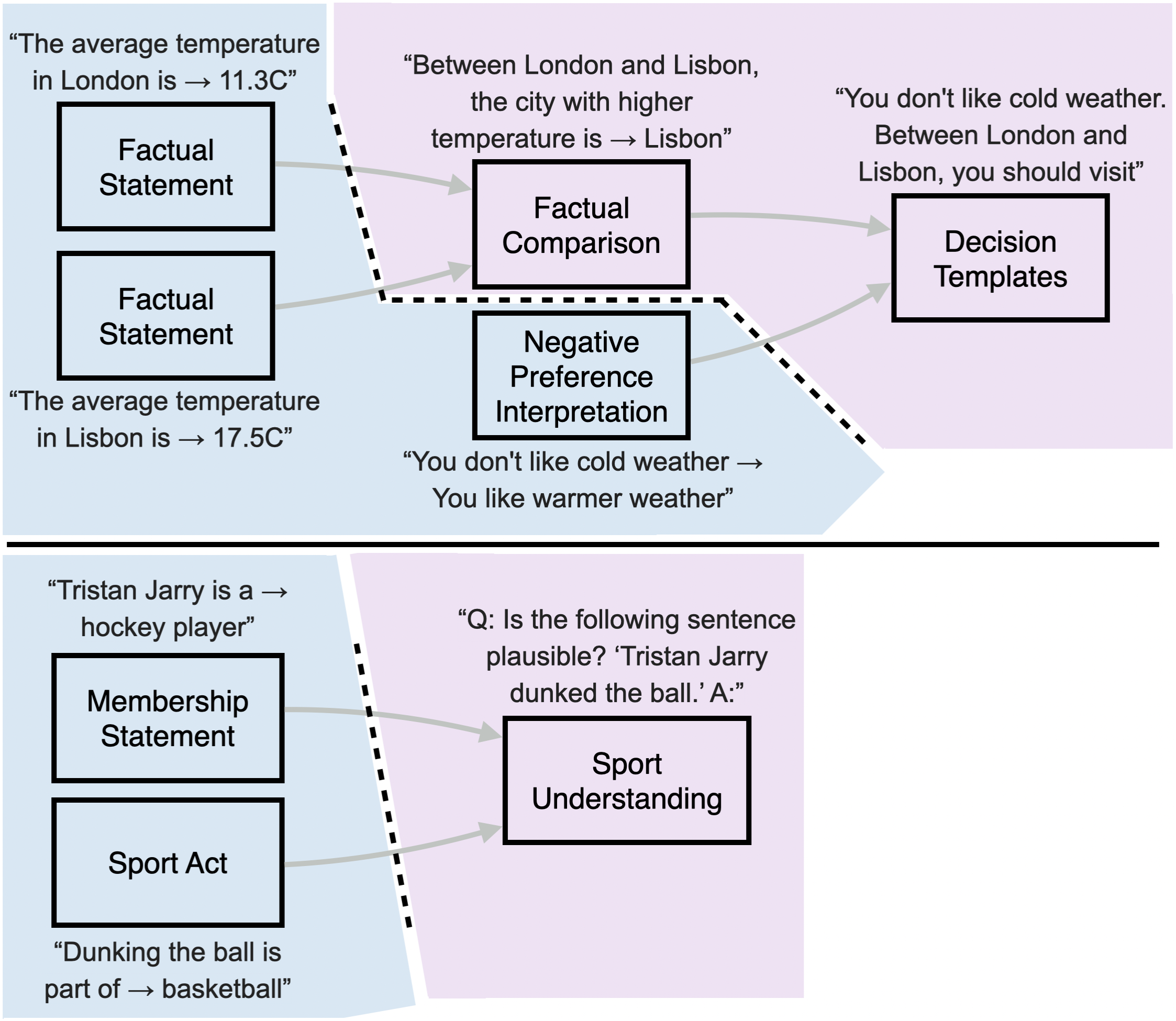}
  \caption{Component tasks involved in a recommendation prompt (above) and in sports understanding (below). In compositional fine-tuning (CFT), component tasks shaded in \textcolor{Cerulean}{light blue} precede those in \textcolor{Thistle}{light purple}.}
  \label{fig:component_tasks}
\end{figure}

Recent work in compositionality in NLP has been mostly limited to semantic parsing and multi-hop reasoning for the purpose of Q\&A \cite{shaw-etal-2021-compositional, 10.1162/tacl_a_00309, min-etal-2019-multi}. However, a series of recent works have proposed generating ``chains of thought'' as a means to expand an LM's ability to reason beyond a single forward pass \cite{wei2022chain, zelikman2022star, nye2021show}. The success of chain of thought approaches suggests broader opportunities to study the use of compositional structure as a means to improve the learning of complex tasks, rather than as a byproduct of end-to-end learning.

Breaking down a complex task into sub-tasks is a ubiquitous construct in human problem-solving. In machine learning, it has inspired curriculum learning (CL) \cite{bengio2009curriculum}, which hypothesizes that a model should start learning from easier concepts and progress to harder ones, as humans do. In this work, we explore the idea of CL through the lens of {\em incremental task complexity}, which is fundamentally different from prior works in NLP centered on incremental example difficulty (e.g., organizing training data by increasing sequence length or decreasing word frequency).

We propose compositional fine-tuning (CFT), a fine-tuning strategy in which sub-tasks are organized as components of a curriculum 
that progressively teaches a target task, as shown visually in Figure \ref{fig:component_tasks}. CFT is novel in two ways: it is a CL approach in NLP that focuses on incremental task complexity instead of incremental example difficulty; and unlike chain of thought prompting, CFT does not depend on huge, pretrained LMs---it relies on smaller, fine-tuned LMs instead. This is advantageous because the largest LMs are hard to access and expensive, and their pretraining data, while vast, still fail to cover a wide range of domains.

We focus on conversational recommendation, which is especially rich in complex tasks \cite{bursztyn-etal-2021-doesnt}. As shown in Figure \ref{fig:component_tasks}, a relatively short recommendation prompt may comprise component tasks as diverse as understanding a user preference---related to pragmatics---and finding an item that correctly matches the semantics of such a preference. Despite this diversity in component tasks, recommendation tasks are still underexplored in the NLP community \cite{10.1145/3383313.3412249, malkiel-etal-2020-recobert, wang2021finetuning}.

We make the following contributions:

\begin{itemize}
    \item We contribute a new schema for generating recommendation datasets, which we instantiate in two domains: world travel and local dining. By design, LMs are more likely to hold prior knowledge about world cities than about local restaurants, making our released dataset challenging to different degrees.
    \item We propose \textbf{compositional fine-tuning (CFT)}: an approach based on decomposing a target task into component tasks, and then fine-tuning smaller LMs on a curriculum of such component tasks. We instantiate CFT in our recommendation tasks as well as the sports understanding task from \cite{wei2022chain}. 
    \item We present experiments\footnote{Data and code fully available at: \url{https://github.com/vbursztyn/compositional-fine-tuning}} showing that CFT consistently outperforms end-to-end learning, with up to 32\% gains in the local dining domain given equal amounts of training data. When compared to chain of thought prompting, we further find that CFT performs equally or better while requiring LMs only 7.4\% of the size (as seen in Table \ref{tab:comparison_chain_of_thought}).
\end{itemize}

\begin{table}[]
\begin{minipage}{.45\textwidth}
\scalebox{0.67}{
\begin{tabular}{|c|c|c|}
\hline
\begin{tabular}[c]{@{}c@{}}Base\\ Model\end{tabular} & Method                                                                                                             & \begin{tabular}[c]{@{}c@{}}Score on\\ Decision\\ Templates\end{tabular} \\ \hline
DaVinci                                              & 8-Shot Prompting                                                                                                   & 0.83 ± 0.08                                                             \\ \hline
DaVinci                                              & 8-Shot Chain of Thought                                                                                            & \textbf{\textcolor{OliveGreen}{0.98 ± 0.02}}                                                             \\ \hline
Curie                                                & 8-Shot Chain of Thought                                                                                            & 0.50 ± 0.12                                                             \\ \hline
Curie                                                & \begin{tabular}[c]{@{}c@{}}CFT on Factual Statements, Factual\\ Comparisons, and Decision Templates\end{tabular} & \textbf{\textcolor{OliveGreen}{0.95 ± 0.01}}                                                             \\ \hline
\end{tabular}}
\end{minipage}

\bigskip

\begin{minipage}{.45\textwidth}
\scalebox{0.67}{
\begin{tabular}{|c|c|c|}
\hline
\begin{tabular}[c]{@{}c@{}}Base\\ Model\end{tabular} & Method                                                                                                             & \begin{tabular}[c]{@{}c@{}}Score on\\ Decision\\ Templates\end{tabular} \\ \hline
DaVinci                                              & 8-Shot Prompting                                                                                                   & 0.54 ± 0.09                                                             \\ \hline
DaVinci                                              & 8-Shot Chain of Thought                                                                                            & \textbf{\textcolor{BrickRed}{0.55 ± 0.07}}                                                             \\ \hline
Curie                                                & 8-Shot Chain of Thought                                                                                            & 0.50 ± 0.06                                                             \\ \hline
Curie                                                & \begin{tabular}[c]{@{}c@{}}CFT on Factual Statements, Factual\\ Comparisons, and Decision Templates\end{tabular} & \textbf{\textcolor{OliveGreen}{0.74 ± 0.05}}                                                             \\ \hline
\end{tabular}}
\caption{Comparison to chain of thought in the world travel domain (above) and local dining (below). CFT performs as well as chain of thought prompting for world cities and 35\% better for local restaurants, with an LM only 7.4\% of the size (13B vs 175B).}
\label{tab:comparison_chain_of_thought}
\end{minipage}
\end{table}

\section{Related Work}
\label{sec:related}

\subsection{Chain of Thought Approaches}

Chain of thought approaches are the most recent stream of research connected to ours \cite{wei2022chain, zelikman2022star, gu2021dream, nye2021show, talmor2020leap, rajani2019explain}. \citet{wei2022chain} recently proposed chain of thought prompting, the idea that very large LMs can do much better at ``system 2 tasks''---tasks that require deeper reasoning skills, such as math problems or symbolic reasoning---if they are given examples in the prompt that explicitly describe the intermediate steps of the task. Although effective in improving accuracy, its dependency on huge, pretrained LMs still limits chain of thought prompting. In contrast, our CFT approach shows similar gains vs end-to-end learning on our tasks, but in a setting with LMs that are more than an order of magnitude smaller.

Among these previous works, we highlight \cite{talmor2020leap} as an attempt to study the effect of factual knowledge injection in LM performance on tasks that involve chaining different facts. In our ablation studies in \S\ref{sec:experiments}, we cover a configuration that is analogous to theirs and show improvements from having an additional component task. 

\subsection{Compositionality in Question Answering}

Many recent works in the Q\&A literature have strived to study compositionality on either a question or system level. At the question level, learning to decompose a question into smaller questions and reasoning over these sub-questions in order to arrive at a final answer (multi-hop reasoning) has been a common goal \cite{khot2020qasc, min-etal-2019-multi, yang-etal-2018-hotpotqa, khashabi-etal-2018-looking}. At the system level, investigating a system's ability to generalize from question types seen during training (e.g., ``Who directed \textit{x}?'') to new, unseen instances of the same type (e.g., ``Who directed Inception?'') has attracted increasing attention \cite{keysers2019measuring}. Further works have explored both problems---multi-hop reasoning and compositional generalization---through the lens of semantic parsing \cite{10.1162/tacl_a_00309, shaw-etal-2021-compositional}.

In contrast, we focus on a new schema of recommendation tasks, where by design the decomposition required to perform the task is not transparent from the question itself but is known {\em a priori} across a variety of domains. This schema allows us to evaluate the effectiveness of a novel CFT approach in two domains, and to compare it against the recent chain of thought prompting approach.

\subsection{Curriculum Learning (CL)}

The seminal work in CL \cite{bengio2009curriculum} included a language modeling experiment in which training data were ordered from most to least frequent based on corpus statistics. Since then, many works in NLP have explored different measures of example difficulty, as simple as sequence length for NLG \cite{rajeswar2017adversarial} and as complex as estimates based on model performance \cite{sachan2016easy, xu-etal-2020-curriculum}. However, such a focus on example difficulty has kept these works distant from the ``shaping hypothesis’’ that inspired \cite{bengio2009curriculum}: the idea that a complex task can be taught by breaking it into a sequence of smaller steps of incremental complexity \cite{krueger2009flexible}. In this work, instead of incremental example difficulty, we explore a different approach to incremental complexity based on organizing training data around component tasks.

To the best of our knowledge, the closest works can be found in the domain of spatial navigation instructions \cite{dan-etal-2021-compositional-data, lake2018generalization}, in which an LM starts with simple block-moving instructions and progresses to compositional ones. However, our work differs in the diversity of our component tasks, in the more extensive experimentation that ensues, and in the applicability of CFT to other similarly diverse domains.

\section{Problem Definition}
\label{sec:problem}

The recommendation task depicted in Figure \ref{fig:component_tasks} takes as input a set of \textit{items} (set $I$) and a set of user preferences (set $P$), such that $Recommend(P, I)$ outputs the item that best matches the user preferences. In its simplest form, we have a pair of items $I = \{i_1$, $i_2$\} and a single preference $P = \{p\}$, such that $Recommend(\{p\},\{i_1, i_2\})$. This form maps naturally to what we call a \textbf{``decision template,''} composed of two sentences: one with a preference (e.g., ``You don't like cold weather.''), and another with a sufficiently different pair of items (e.g., ``Between London and Lisbon, you should visit'' $\rightarrow$ Lisbon). We use the term ``decision'' because $Recommend(P, I)$ can be considered an instance of a decision task where $I$ represents options and $P$ expresses the criteria to be applied.

Breaking down $Recommend(\{p\},\{i_1, i_2\})$ into component tasks, the first task consists of comparing two items along a given attribute. This can be defined as $Compare(a, o, \{i_1, i_2\})$ that takes as input an attribute $a$ (e.g., temperature), an order $o$ (e.g., higher), and the two items, and then outputs the item that satisfies the comparison. We call this task a \textbf{``factual comparison''} (e.g., ``Between London and Lisbon, the city with warmer weather is'' $\rightarrow$ Lisbon), which is further decomposed into \textbf{``factual statements''} that simply enunciate the attribute value of an item (e.g., ``The average temperature in Lisbon is'' $\rightarrow$ 17.5C).

With that, a domain $D$ can be formalized as $D = (I^{full},A)$ where $I^{full}$ is the full set of items and $A$ the set of attributes. Considering the world travel domain, for example, $I^{full}$ may represent a list of well-known cities and $A=\{temperature, population\}$ the average temperature and total population, respectively. We instantiate this schema in our experiments in \S\ref{sec:experiments}, but it can be used to generate new recommendation datasets or repurposed for other decision tasks.

\subsection{A Challenging Task for Pretrained LMs}
\label{sub:PLMs}

Even state-of-the-art LMs such as GPT-3 \cite{brown2020language} struggle at this recommendation task, as evidenced by experiments fully described in \S\ref{sec:experiments}. As shown in Table \ref{tab:comparison_chain_of_thought}, 175B parameter DaVinci in 8-shot mode can accurately recommend 83\% of test cases in the world travel domain, but only 55\% in the local dining domain, which cannot be improved with chain of thought prompting. As shown in Tables \ref{tab:exp1_cities} and \ref{tab:exp1_restaurants}, performance is very low with 13B parameter Curie in 0-shot mode: only 6\% of test cases lead to correct recommendations in the world travel domain, and only 18\% in the local dining domain. It is with this challenge in mind that we propose compositional fine-tuning (CFT).

\section{Compositional Fine-Tuning (CFT)}
\label{sec:CFT}

CFT consists of three sequential steps: \textbf{Decompose}, where we break the complex task into component tasks; \textbf{Demonstrate}, where we generate examples for each of these component tasks; and \textbf{Fine-Tune}, where we organize the training data according to task-level compositionality.

\subsection{Decompose}

For the Decompose step, Figure \ref{fig:component_tasks} and \S\ref{sec:problem} establish the component tasks behind decision templates. In this work, the decomposition is performed manually in order to evaluate whether using compositional structure during fine-tuning can potentially improve the learning of complex tasks. This assumption is similar in spirit to the step-by-step ``exemplars'' manually provided in chain of thought prompting \cite{wei2022chain}. In line with their findings, we believe that the confirmation of our hypothesis helps to motivate further research in automating this step.

\subsection{Demonstrate}
\label{sub:demonstrate}

Once we have a diagram with component tasks, we need to demonstrate them, preferably with some degree of natural language variation. In our recommendation dataset, we implement a single factual comparison (e.g., comparing London and Lisbon with $a =$ temperature and $o =$ higher) using two different phrasings, and the corresponding decision template using eight different phrasings.

For factual comparisons, the first phrasing {\em directly} refers to the attribute value (e.g., ``Between London and Lisbon, the city with higher average temperature is'' $\rightarrow$ Lisbon), and the second phrasing {\em indirectly} refers to the same attribute value (e.g., ``Between London and Lisbon, the city with warmer weather is'' $\rightarrow$ Lisbon).

For decision templates, following \cite{bursztyn-etal-2021-doesnt}, each possible $a$ and $o$ combination (i.e., each preference) is phrased in either a positive form (e.g., ``You like warmer weather.'') or a negative form (e.g., ``You don't like cold weather.''). Additionally, each of these two phrasings can be rephrased in the first- or third-person (``Someone...''), as well as in a subjunctive form (e.g., ``You are looking for a city with warmer weather. If I were you, I would visit'').

Completing our setting, as seen in \S\ref{sec:problem}, factual statements are represented by a single phrasing that simply enunciates an attribute value. Therefore, with $|A|=2$, each pair of items yields four factual statements, eight factual comparisons, and 32 decision templates.\footnote{Fully available at: \url{https://bit.ly/3xeP8E1}}

\subsection{Fine-Tune}

Once all these phrasings are populated with item pairs from one of our two domains, we are done generating our training data. For the Fine-Tune step, we organize such data according to component tasks' dependencies. As seen in Figure \ref{fig:component_tasks}, there are tasks that do not depend on any other (in \textcolor{Cerulean}{light blue}), while there are tasks that do (\textcolor{Thistle}{light purple}).

From left to right, we consider each colored layer a phase in our curriculum: the first phase includes data for factual statements and negative preference interpretations; and the second phase includes factual comparisons and decision templates. As explained in \S\ref{sec:exp2}, negative preference interpretations are a small component task that is useful when decision templates are partially seen during training.


Within each phase, we find empirical benefits in shuffling training data. We put forward two potential explanations for that. First, in earlier phases, shuffling ensures that all component tasks included in a phase are equally learned by the end of it, helping in the next one. Second, in later phases, shuffling should also help training to converge because these later tasks are increasingly similar to the target task.

\section{Experiments}
\label{sec:experiments}

Considering our problem and CFT, we pose the following questions:

\begin{itemize}
    \item \textbf{RQ1}: How does CFT compare with end-to-end learning?
    \item \textbf{RQ2}: How does CFT compare with chain of thought prompting?
\end{itemize}

We conduct four experiments to answer RQs 1 and 2, leveraging data from two domains: world travel, and local dining. World travel represents a less challenging scenario, considering that the LM is more likely to have prior knowledge about world cities and their various attributes. Local dining, conversely, represents a more challenging scenario, as the LM is less likely to exhibit any prior knowledge. Our experiments are based on GPT-3's Curie model (13B parameters), which was the largest LM available for fine-tuning at that time.\footnote{\url{https://beta.openai.com/docs/engines}}

\subsection{Data}

Each domain comprises two attributes. For world cities, we have $A = \{temperature, population\}$ where average city temperatures are obtained from Wikipedia\footnote{\url{https://en.wikipedia.org/wiki/List_of_cities_by_average_temperature}} and city populations from SimpleMaps 2019.\footnote{\url{https://simplemaps.com/data/world-cities}} After merging items from both sources, we end with 347 well-known cities (>50k inhabitans) from around the globe, such that $D_{c}=(I^{full}_{c}, \{temperature, population\})$ and $|I^{full}_{c}| = 347$. For local restaurants, we randomly sample 240 restaurants from the city with most restaurants in the Yelp dataset\footnote{\url{https://www.yelp.com/dataset}}, Toronto. We have $A = \{price, distance\}$ where restaurant prices are obtained from Yelp and distances to a hypothetical location are randomly generated, thus limiting the LM's access to prior knowledge in this scenario. With that, we have $D_{r}=(I^{full}_{r}, \{price, distance\})$ and $|I^{full}_{r}| = 240$.

In terms of component tasks, we have 694 factual statements for the cities domain and 480 for restaurants, covering two attributes per item. Whenever factual statements are provided in CFT, they always cover $I^{full}$ entirely in order to give the LM full knowledge of the attribute values.

However, for factual comparisons and decision templates, we wish to evaluate the LM's ability to generalize to cities and restaurants not seen in such statements during training. Therefore, we split $I^{full}$ between training and test items before we generate item pairs. We keep only 30\% of $I^{full}$ for training, and we sample from the remaining 70\% when testing a fine-tuned LM. This way, cities and restaurants used at test time are only seen during training in factual statements, {\em never} in factual comparisons or decision templates.

When generating examples from these itemsets, we enforce minimum differences in attribute values: for pairs of cities, a 10C difference in temperature and a 2.5M difference in population; and for pairs of restaurants, a 1 dollar-sign difference in price and a 3 mile difference in distance. Factual comparisons and decision templates are only populated with item pairs that exhibit at least these differences in attribute values.

When applying these rules to the training items, we end with roughly 1,970 pairs of cities and 2,320 pairs of restaurants. In combination with the phrasings in \S\ref{sub:demonstrate}, we have roughly 15.8k factual comparisons for cities and 18.5k for restaurants; and 63k decision templates for cities and 74.2k for restaurants. To make sure that factual comparisons and decision templates are represented by similar amounts of training data, we sample decision templates until the number of tokens match that of factual comparisons.

Lastly, across all factual comparisons and decision templates, we flip the order of the items (e.g., London and Lisbon) with a 50\% chance so that the LM cannot use position as a short-cut for the answer. Our data is made fully available to the research community at \url{https://github.com/vbursztyn/compositional-fine-tuning}

\subsection{Evaluation}

Once we fine-tune Curie on a given CFT configuration, we may evaluate the model on a task from the second phase---either factual comparisons or decision templates---in a given domain. We generate examples by applying the same rules seen in the generation of training data, but now applied to the held-out test items. When evaluating factual comparisons, we report the average performance on 1.6k test cases (200 examples per phrasing, times eight phrasings); and when evaluating decision templates, we report the average performance on 6.4k test cases (200 times 32 phrasings).

A single test case is evaluated by generating the top 5 predictions with greedy decoding. If the answer is more likely than the wrong candidate, then this test case score is 1; otherwise, it is 0.

\subsection{Experiment 1: The Role of Components}

In our first experiment, we want to answer RQ1 by examining the role of component tasks in the learning of our complex task. To this end, we focus on the deepest dependencies of decision templates, i.e., factual comparisons and factual statements. We ablate each of these component tasks in different CFT configurations while measuring model performance on decision templates. Although performance on decision templates is our primary end-point, we secondarily measure performance on factual comparisons. Tables \ref{tab:exp1_cities} and \ref{tab:exp1_restaurants} show the results of each CFT configuration for world cities and local restaurants, respectively.

\begin{table}[]
\begin{minipage}{.45\textwidth}
\scalebox{0.63}{
\begin{tabular}{|ccc|cc|}
\hline
\multicolumn{3}{|c|}{Model fine-tuned on}                                                                                                                                                                                              & \multicolumn{2}{c|}{Average score on}                                                                                                             \\ \hline
\multicolumn{1}{|c|}{\begin{tabular}[c]{@{}c@{}}Factual\\ Statements\end{tabular}} & \multicolumn{1}{c|}{\begin{tabular}[c]{@{}c@{}}Factual\\ Comparisons\end{tabular}} & \begin{tabular}[c]{@{}c@{}}Decision\\ Templates\end{tabular} & \multicolumn{1}{c|}{\begin{tabular}[c]{@{}c@{}}Factual\\ Comparisons\end{tabular}} & \begin{tabular}[c]{@{}c@{}}Decision\\ Templates\end{tabular} \\ \hline
\multicolumn{1}{|c|}{{\color[HTML]{C0C0C0} No}}                                                          & \multicolumn{1}{c|}{{\color[HTML]{C0C0C0} No}}                                     & {\color[HTML]{C0C0C0} No}                                    & \multicolumn{1}{c|}{0.16 ± 0.06}                                                 & 0.06 ± 0.07
\\ \hline
\multicolumn{1}{|c|}{Yes}                                                          & \multicolumn{1}{c|}{{\color[HTML]{C0C0C0} No}}                                     & {\color[HTML]{C0C0C0} No}                                    & \multicolumn{1}{c|}{0.11 ± 0.07}                                                 & 0.27 ± 0.15                                                \\ \hline
\multicolumn{1}{|c|}{{\color[HTML]{C0C0C0} No}}                                    & \multicolumn{1}{c|}{Yes}                                                           & {\color[HTML]{C0C0C0} No}                                    & \multicolumn{1}{c|}{0.90 ± 0.02}                                                 & 0.54 ± 0.17                                                \\ \hline
\multicolumn{1}{|c|}{{\color[HTML]{C0C0C0} No}}                                    & \multicolumn{1}{c|}{{\color[HTML]{C0C0C0} No}}                                     & Yes                                                          & \multicolumn{1}{c|}{0.74 ± 0.16}                                                 & 0.89 ± 0.04                                                \\ \hline
\multicolumn{1}{|c|}{Yes}                                                          & \multicolumn{1}{c|}{Yes}                                                           & {\color[HTML]{C0C0C0} No}                                    & \multicolumn{1}{c|}{0.95 ± 0.02}                                                 & 0.63 ± 0.18                                                \\ \hline
\multicolumn{1}{|c|}{Yes}                                                          & \multicolumn{1}{c|}{{\color[HTML]{C0C0C0} No}}                                     & Yes                                                          & \multicolumn{1}{c|}{0.78 ± 0.22}                                                 & 0.92 ± 0.02                                                \\ \hline
\multicolumn{1}{|c|}{{\color[HTML]{C0C0C0} No}}                                    & \multicolumn{1}{c|}{Yes}                                                           & Yes                                                          & \multicolumn{1}{c|}{0.89 ± 0.03}                                                 & 0.88 ± 0.03                                                \\ \hline
\multicolumn{1}{|c|}{Yes}                                                          & \multicolumn{1}{c|}{Yes}                                                           & Yes                                                          & \multicolumn{1}{c|}{0.96 ± 0.01}                                                 & \textbf{0.95 ± 0.01}                                       \\ \hline
\end{tabular}}
\caption{Experiment 1 in the world travel domain. CFT with factual statements or factual comparisons consistently increases performance. The best configuration includes all tasks (row \#8, in boldface).}
\label{tab:exp1_cities}
\end{minipage}

\begin{minipage}{.45\textwidth}
\scalebox{0.63}{
\begin{tabular}{|ccc|cc|}
\hline
\multicolumn{3}{|c|}{Model fine-tuned on}                                                                                                                                                                                              & \multicolumn{2}{c|}{Average score on}                                                                                                             \\ \hline
\multicolumn{1}{|c|}{\begin{tabular}[c]{@{}c@{}}Factual\\ Statements\end{tabular}} & \multicolumn{1}{c|}{\begin{tabular}[c]{@{}c@{}}Factual\\ Comparisons\end{tabular}} & \begin{tabular}[c]{@{}c@{}}Decision\\ Templates\end{tabular} & \multicolumn{1}{c|}{\begin{tabular}[c]{@{}c@{}}Factual\\ Comparisons\end{tabular}} & \begin{tabular}[c]{@{}c@{}}Decision\\ Templates\end{tabular} \\ \hline
\multicolumn{1}{|c|}{{\color[HTML]{C0C0C0} No}}                                                          & \multicolumn{1}{c|}{{\color[HTML]{C0C0C0} No}}                                     & {\color[HTML]{C0C0C0} No}                                    & \multicolumn{1}{c|}{0.16 ± 0.04}                                                 & 0.18 ± 0.05
\\ \hline
\multicolumn{1}{|c|}{Yes}                                                          & \multicolumn{1}{c|}{{\color[HTML]{C0C0C0} No}}                                     & {\color[HTML]{C0C0C0} No}                                    & \multicolumn{1}{c|}{0.00 ± 0.00}                                                 & 0.13 ± 0.06                                                \\ \hline
\multicolumn{1}{|c|}{{\color[HTML]{C0C0C0} No}}                                    & \multicolumn{1}{c|}{Yes}                                                           & {\color[HTML]{C0C0C0} No}                                    & \multicolumn{1}{c|}{0.52 ± 0.11}                                                 & 0.51 ± 0.10                                                \\ \hline
\multicolumn{1}{|c|}{{\color[HTML]{C0C0C0} No}}                                    & \multicolumn{1}{c|}{{\color[HTML]{C0C0C0} No}}                                     & Yes                                                          & \multicolumn{1}{c|}{0.50 ± 0.06}                                                 & 0.52 ± 0.07                                                \\ \hline
\multicolumn{1}{|c|}{Yes}                                                          & \multicolumn{1}{c|}{Yes}                                                           & {\color[HTML]{C0C0C0} No}                                    & \multicolumn{1}{c|}{0.66 ± 0.13}                                                 & 0.54 ± 0.10                                                \\ \hline
\multicolumn{1}{|c|}{Yes}                                                          & \multicolumn{1}{c|}{{\color[HTML]{C0C0C0} No}}                                     & Yes                                                          & \multicolumn{1}{c|}{0.50 ± 0.05}                                                 & 0.55 ± 0.04                                                \\ \hline
\multicolumn{1}{|c|}{{\color[HTML]{C0C0C0} No}}                                    & \multicolumn{1}{c|}{Yes}                                                           & Yes                                                          & \multicolumn{1}{c|}{0.53 ± 0.12}                                                 & 0.53 ± 0.10                                                \\ \hline
\multicolumn{1}{|c|}{Yes}                                                          & \multicolumn{1}{c|}{Yes}                                                           & Yes                                                          & \multicolumn{1}{c|}{0.75 ± 0.05}                                                 & \textbf{0.74 ± 0.05}                                       \\ \hline
\end{tabular}}
\caption{Experiment 1 in the local dining domain. The best configuration, again with all tasks, outperforms the second best (row \#6) by 35\%.}
\label{tab:exp1_restaurants}
\end{minipage}
\end{table}

We can see that factual statements consistently improve performance: on Table \ref{tab:exp1_cities}, they improve performance by 3-17\%, including an 8\% improvement of row \#8 (the best configuration) relative to row \#7. On Table \ref{tab:exp1_restaurants}, they improve performance by 5-40\%, with maximum improvement on row \#8 (again, the best configuration) over row \#7. This component task has a small footprint---694 factual statements for cities and 480 for restaurants---and is the most likely one to be contemplated in end-to-end learning schemes (e.g., when knowledge bases are included during training).

We can also see that factual comparisons monotonically increase performance: on Table \ref{tab:exp1_cities}, although there is no change from row \#7 to row \#4, they improve row \#8 by 3\% over row \#6. On Table \ref{tab:exp1_restaurants}, although again there is no change from row \#7 to row \#4, they improve row \#8 by 35\% over row \#6. Therefore, in the best configuration, the effect of factual comparisons is comparable to that of factual statements (35\% vs 40\%). Scores on factual comparisons also suggest that the learning of both tasks in the second phase of CFT is indeed synergistic.

Interestingly, the second best configuration (row \#6) represents an end-to-end learning scheme with access to factual knowledge, which is similar to configurations studied by \cite{talmor2020leap}. However, because factual comparisons and decision templates were designed to have the same number of tokens in our CFT configurations, row \#8 has access to almost two times as much training data as row \#6.

For this reason, we run a follow-up experiment to test if the performance gains are indeed explained by the presence of more components, and not by access to more data. We increase the number of decision templates in row \#6 until we have equal amounts of training data.

\begin{table}[]
\begin{minipage}{.48\textwidth}
\scalebox{0.55}{
\begin{tabular}{|c|ccc|cc|}
\hline
                                                                                 & \multicolumn{3}{c|}{Model fine-tuned on}                                                                                                                                                                                              & \multicolumn{2}{c|}{Average score on}                                                                                                             \\ \cline{2-6} 
\multirow{-2}{*}{\begin{tabular}[c]{@{}c@{}}Total \#\\of tokens\end{tabular}} & \multicolumn{1}{c|}{\begin{tabular}[c]{@{}c@{}}Factual\\ Statements\end{tabular}} & \multicolumn{1}{c|}{\begin{tabular}[c]{@{}c@{}}Factual\\ Comparisons\end{tabular}} & \begin{tabular}[c]{@{}c@{}}Decision\\ Templates\end{tabular} & \multicolumn{1}{c|}{\begin{tabular}[c]{@{}c@{}}Factual\\ Comparisons\end{tabular}} & \begin{tabular}[c]{@{}c@{}}Decision\\ Templates\end{tabular} \\ \hline
186,413                                                                          & \multicolumn{1}{c|}{Yes}                                                          & \multicolumn{1}{c|}{{\color[HTML]{C0C0C0} No}}                                     & Yes                                                          & \multicolumn{1}{c|}{0.78 ± 0.22}                                                 & 0.92 ± 0.02                                                \\ \hline
367,144                                                                          & \multicolumn{1}{c|}{Yes}                                                          & \multicolumn{1}{c|}{{\color[HTML]{C0C0C0} No}}                                     & Yes                                                          & \multicolumn{1}{c|}{0.75 ± 0.26}                                                 & 0.93 ± 0.02                                                \\ \hline
367,157                                                                          & \multicolumn{1}{c|}{Yes}                                                          & \multicolumn{1}{c|}{Yes}                                                           & Yes                                                          & \multicolumn{1}{c|}{0.96 ± 0.01}                                                 & \textbf{0.95 ± 0.01}                                       \\ \hline
\end{tabular}}
\caption{CFT vs end-to-end learning (plus facts) with equal amounts of training data for world cities. The gap practically does not change.}
\label{tab:exp1b_cities}
\end{minipage}
\end{table}

\begin{table}[]
\begin{minipage}{.48\textwidth}
\scalebox{0.55}{
\begin{tabular}{|c|ccc|cc|}
\hline
                                                                               & \multicolumn{3}{c|}{Model fine-tuned on}                                                                                                                                                                                              & \multicolumn{2}{c|}{Average score on}                                                                                                             \\ \cline{2-6} 
\multirow{-2}{*}{\begin{tabular}[c]{@{}c@{}}Total \#\\ of tokens\end{tabular}} & \multicolumn{1}{c|}{\begin{tabular}[c]{@{}c@{}}Factual\\ Statements\end{tabular}} & \multicolumn{1}{c|}{\begin{tabular}[c]{@{}c@{}}Factual\\ Comparisons\end{tabular}} & \begin{tabular}[c]{@{}c@{}}Decision\\ Templates\end{tabular} & \multicolumn{1}{c|}{\begin{tabular}[c]{@{}c@{}}Factual\\ Comparisons\end{tabular}} & \begin{tabular}[c]{@{}c@{}}Decision\\ Templates\end{tabular} \\ \hline
293,967                                                                        & \multicolumn{1}{c|}{Yes}                                                          & \multicolumn{1}{c|}{{\color[HTML]{C0C0C0} No}}                                     & Yes                                                          & \multicolumn{1}{c|}{0.50 ± 0.05}                                                 & 0.55 ± 0.04                                                \\ \hline
581,370                                                                        & \multicolumn{1}{c|}{{\color[HTML]{333333} Yes}}                                   & \multicolumn{1}{c|}{{\color[HTML]{C0C0C0} No}}                                     & Yes                                                          & \multicolumn{1}{c|}{0.50 ± 0.10}                                                 & 0.56 ± 0.04                                                \\ \hline
581,379                                                                        & \multicolumn{1}{c|}{Yes}                                                          & \multicolumn{1}{c|}{Yes}                                                           & Yes                                                          & \multicolumn{1}{c|}{0.75 ± 0.05}                                                 & \textbf{0.74 ± 0.05}                                       \\ \hline
\end{tabular}}
\caption{CFT vs end-to-end learning (plus facts) with equal amounts of training data for local restaurants. Again, the gap practically does not change.}
\label{tab:exp1b_restaurants}
\end{minipage}
\end{table}

On Tables \ref{tab:exp1b_cities} and \ref{tab:exp1b_restaurants}, we can see how the quantity of training data does not explain the performance difference. With equal amounts of training data, \textbf{our CFT configuration with more component tasks consistently outperforms end-to-end learning} with factual knowledge: by 2\% for world cities, and up to 32\% for local restaurants. Importantly, CFT yields substantial improvements in the more challenging scenario where the LM has less prior knowledge on items, thus a performance that is further from the upper bound.

\subsection{Experiment 2: Attribute Transfer}
\label{sec:exp2}

In our second experiment, we continue to address the question: Are more component tasks better for CFT? To complement Experiment 1, we split the original decision templates data into two folds, one for each attribute, and we ablate these folds in each domain while measuring model performance on the entire set of decision templates.

\begin{table*}[ht]
\centering
\scalebox{0.63}{
\begin{tabular}{|ccccc|cc|}
\hline
\multicolumn{5}{|c|}{Model fine-tuned on}                                                                                                                                                                                                                                                                                                                                                                                                                    & \multicolumn{2}{c|}{Average score on}                                                                                                             \\ \hline
\multicolumn{1}{|c|}{\begin{tabular}[c]{@{}c@{}}Factual\\ Statements\end{tabular}} & \multicolumn{1}{c|}{\begin{tabular}[c]{@{}c@{}}Factual\\ Comparisons\end{tabular}} & \multicolumn{1}{c|}{\begin{tabular}[c]{@{}c@{}}Decision\\ Templates\\ (Weather)\end{tabular}} & \multicolumn{1}{c|}{\begin{tabular}[c]{@{}c@{}}Decision\\ Templates\\ (Population)\end{tabular}} & \begin{tabular}[c]{@{}c@{}}Negative\\ Preference\\ Interpretations\end{tabular} & \multicolumn{1}{c|}{\begin{tabular}[c]{@{}c@{}}Factual\\ Comparisons\end{tabular}} & \begin{tabular}[c]{@{}c@{}}Decision\\ Templates\end{tabular} \\ \hline
\multicolumn{1}{|c|}{}                                                             & \multicolumn{1}{c|}{}                                                              & \multicolumn{1}{c|}{{\color[HTML]{C0C0C0} }}                                                  & \multicolumn{1}{c|}{{\color[HTML]{333333} }}                                                     & {\color[HTML]{C0C0C0} No}                                                       & \multicolumn{1}{c|}{0.94 ± 0.01}                                                 & 0.84 ± 0.19                                                \\ \cline{5-7} 
\multicolumn{1}{|c|}{}                                                             & \multicolumn{1}{c|}{}                                                              & \multicolumn{1}{c|}{\multirow{-2}{*}{{\color[HTML]{C0C0C0} No}}}                              & \multicolumn{1}{c|}{\multirow{-2}{*}{{\color[HTML]{333333} Yes}}}                                & {\color[HTML]{333333} Yes}                                                      & \multicolumn{1}{c|}{0.95 ± 0.01}                                                 & \textbf{0.89 ± 0.10}                                       \\ \cline{3-7} 
\multicolumn{1}{|c|}{}                                                             & \multicolumn{1}{c|}{}                                                              & \multicolumn{1}{c|}{{\color[HTML]{333333} }}                                                  & \multicolumn{1}{c|}{{\color[HTML]{C0C0C0} }}                                                     & {\color[HTML]{C0C0C0} No}                                                       & \multicolumn{1}{c|}{0.95 ± 0.01}                                                 & 0.88 ± 0.12                                                \\ \cline{5-7} 
\multicolumn{1}{|c|}{\multirow{-4}{*}{Yes}}                                        & \multicolumn{1}{c|}{\multirow{-4}{*}{Yes}}                                         & \multicolumn{1}{c|}{\multirow{-2}{*}{{\color[HTML]{333333} Yes}}}                             & \multicolumn{1}{c|}{\multirow{-2}{*}{{\color[HTML]{C0C0C0} No}}}                                 & Yes                                                                             & \multicolumn{1}{c|}{0.96 ± 0.01}                                                 & \textbf{0.90 ± 0.14}                                       \\ \hline
\end{tabular}}
\caption{Experiment 2 for world cities. Adding only 12 negative preference interpretations improves performance by 2-6\% on the two folds.}
\label{tab:exp2_cities}

\scalebox{0.63}{
\begin{tabular}{|ccccc|cc|}
\hline
\multicolumn{5}{|c|}{Model fine-tuned on}                                                                                                                                                                                                                                                                                                                                                                                                                & \multicolumn{2}{c|}{Average score on}                                                                                                             \\ \hline
\multicolumn{1}{|c|}{\begin{tabular}[c]{@{}c@{}}Factual\\ Statements\end{tabular}} & \multicolumn{1}{c|}{\begin{tabular}[c]{@{}c@{}}Factual\\ Comparisons\end{tabular}} & \multicolumn{1}{c|}{\begin{tabular}[c]{@{}c@{}}Decision\\ Templates\\ (Price)\end{tabular}} & \multicolumn{1}{c|}{\begin{tabular}[c]{@{}c@{}}Decision\\ Templates\\ (Distance)\end{tabular}} & \begin{tabular}[c]{@{}c@{}}Negative\\ Preference\\ Interpretations\end{tabular} & \multicolumn{1}{c|}{\begin{tabular}[c]{@{}c@{}}Factual\\ Comparisons\end{tabular}} & \begin{tabular}[c]{@{}c@{}}Decision\\ Templates\end{tabular} \\ \hline
\multicolumn{1}{|c|}{}                                                             & \multicolumn{1}{c|}{}                                                              & \multicolumn{1}{c|}{{\color[HTML]{C0C0C0} }}                                                & \multicolumn{1}{c|}{{\color[HTML]{333333} }}                                                   & {\color[HTML]{C0C0C0} No}                                                       & \multicolumn{1}{c|}{0.64 ± 0.13}                                                 & 0.56 ± 0.06                                                \\ \cline{5-7} 
\multicolumn{1}{|c|}{}                                                             & \multicolumn{1}{c|}{}                                                              & \multicolumn{1}{c|}{\multirow{-2}{*}{{\color[HTML]{C0C0C0} No}}}                            & \multicolumn{1}{c|}{\multirow{-2}{*}{{\color[HTML]{333333} Yes}}}                              & {\color[HTML]{333333} Yes}                                                      & \multicolumn{1}{c|}{0.70 ± 0.08}                                                 & \textbf{0.65 ± 0.05}                                       \\ \cline{3-7} 
\multicolumn{1}{|c|}{}                                                             & \multicolumn{1}{c|}{}                                                              & \multicolumn{1}{c|}{{\color[HTML]{333333} }}                                                & \multicolumn{1}{c|}{{\color[HTML]{C0C0C0} }}                                                   & {\color[HTML]{C0C0C0} No}                                                       & \multicolumn{1}{c|}{0.68 ± 0.13}                                                 & 0.67 ± 0.17                                                \\ \cline{5-7} 
\multicolumn{1}{|c|}{\multirow{-4}{*}{Yes}}                                        & \multicolumn{1}{c|}{\multirow{-4}{*}{Yes}}                                         & \multicolumn{1}{c|}{\multirow{-2}{*}{{\color[HTML]{333333} Yes}}}                           & \multicolumn{1}{c|}{\multirow{-2}{*}{{\color[HTML]{C0C0C0} No}}}                               & Yes                                                                             & \multicolumn{1}{c|}{0.67 ± 0.17}                                                 & \textbf{0.69 ± 0.16}                                       \\ \hline
\end{tabular}}
\caption{Experiment 2 for local restaurants. Again, adding only 12 negative preference interpretations improves performance by 3-16\% on the two folds.}
\label{tab:exp2_restaurants}
\end{table*}

On Tables \ref{tab:exp2_cities} and \ref{tab:exp2_restaurants}, when analyzing the configurations on rows \#1 and \#3, we notice that learning is partially transferred to the unseen attribute, with performance drops of 7-24\% relative to rows \#8 of Tables \ref{tab:exp1_cities} and \ref{tab:exp1_restaurants}. We also notice that unseen preferences phrased in the negative form (e.g., ``You don't like cold weather.'') are the biggest source of error. Therefore, we add one extra component task to these CFT configurations: negative preference interpretations.\footnote{Fully available at: \url{https://bit.ly/3O0WIce}}

As seen in Figure \ref{fig:component_tasks}, these interpretations simply teach the LM to interpret negations (e.g., ``You don't like cold weather'' $\rightarrow$ ``You like warmer weather''), consisting of only 12 statements for each domain---a tiny footprint. Interestingly, this small component task indeed improves performance across all configurations: 2-6\% for world cities, and 3-16\% for local restaurants. Analyzing {\em exclusively} the decision templates containing negations, performance improves by an average of 9\% for cities and 15\% for restaurants.

\subsection{Experiments 3 \& 4: Comparison to Chain of Thought Prompting}

In our two final experiments, we want to answer RQ2 by comparing CFT with chain of thought prompting. We do this from two perspectives: first, from the viewpoint of the recommendation tasks introduced in this work; and second, from the viewpoint of sports understanding, a commonsense task studied by \cite{wei2022chain}.

\subsubsection{Recommendation Tasks}
\label{sec:cot_recommendation}

We instantiate chain of thought prompting in our two domains as described in \cite{wei2022chain}, with $k = 8$ per their code. For each domain, we manually construct 8 ``exemplars'' using item pairs that only exist in the training set. Each exemplar includes relevant factual statements, how these are used in a factual comparison, and how this is used to answer the overarching decision template.\footnote{Prompts available at: \url{https://bit.ly/3rDiwS6}} As recommended by \cite{wei2022chain}, we leverage pretrained DaVinci (175B parameters), which is the largest LM we have access to; but we also test pretrained Curie (13B parameters), which is the base model for all our CFT runs. Finally, to isolate the effect of chain of thought, we test DaVinci with regular 8-shot prompting. Due to the much higher costs of 8-shot chain of thought prompting with DaVinci, in these runs we reduced the sample size to 100 test cases per phrasing (3.2k test cases in total). Results can be seen on Table \ref{tab:comparison_chain_of_thought}.

Interestingly, for world cities, chain of thought prompting with pretrained DaVinci can answer almost all test cases (98\% of them), which is not too far ahead of CFT using Curie (95\%). Pretrained DaVinci with regular 8-shot prompting performs substantially worse (83\%), which shows how both chain of thought prompting and CFT are more effective in this scenario. However, chain of thought prompting with pretrained Curie performs as low as random chance (50\%). This suggests that \textbf{CFT is capable of similar performance while requiring an LM only 7.4\% of the size} (13B vs 175B).

For local restaurants, results are even more favorable for CFT. All approaches based on pretrained LMs struggle in this more challenging domain, performing only slightly above chance on the price attribute (up to 65\%). CFT is the only approach capable of answering 74\% of all test cases, which shows a fundamental limitation of chain of thought prompting when faced with domains where facts are not as easily accessible.

\subsubsection{Sports Understanding}

Next, we instantiate CFT in the sports understanding task from \cite{wei2022chain}, which consists in determining if a sentence mentioning a certain well-known sport player performing a certain sport act is plausible. Component tasks can be seen in Figure \ref{fig:component_tasks}. We note that this task is very similar in structure to the ``hypernymy'' and ``meronymy'' tasks from \cite{talmor2020leap}, thus also representing a large category of inferential tasks.

To gather data for the Demonstrate step, we resort to a similar strategy to \cite{zelikman2022star}: using the generated chains of thought in \cite{wei2022chain}, we filter all the explanations that lead to a correct answer. From these 815 examples (originally 980, given their accuracy of 83\% on the task), we parse 390 unique membership statements and 182 unique sport acts. Our CFT configuration includes all membership statements and sport acts, but only 50\% of question-answer pairs in a 2-fold cross-validation scheme. Performance on the two folds are 95.83\% and 95.57\%. Therefore, like in \S\ref{sec:cot_recommendation}, this again suggests that CFT is capable of similar performance to chain of thought promping while requiring an LM only 7.4\% of the size.

\section{Discussion}
\label{sec:discussion}

Across our experiments, both in our released dataset and in sports understanding, we found consistent evidence that LMs may benefit from compositional structure when learning a complex task. Although we obtain improvements from {\em three very different} types of component tasks---factual statements, factual comparisons, and negative preference interpretations---standard end-to-end learning schemes tend to overlook the explicit use of compositional structure or focus only on factual knowledge. We hope to encourage further research in other principled, task-agnostic methods for leveraging compositional structure in LM fine-tuning.

Compared to chain of thought prompting, methods based on fine-tuning have at least two advantages. First, 100+B parameter LMs are hard to access and expensive. When using DaVinci with 8-shot chain of thought prompting, each of our examples costs USD 7.5 cents,\footnote{Two requests required, with 625 prompt tokens each. Querying DaVinci currently costs USD 6 cents per 1k tokens.} which is roughly {\em 50 times more expensive} than fine-tuning Curie with CFT. Second, many domains are not within pretraining data (e.g., due to proprietary data), so it is necessary to consider fine-tuning methods that inject custom data {\em and} preserve the LM's ability to chain thought. This limitation of strictly prompt-based methods has been recently noted by \cite{zhou2022least}, and we emphasize it in light of our results in the local dining domain. 

While CFT certainly requires more data than chain of thought prompting, interestingly, we found it to be remarkably more efficient w.r.t model size. Works leading up to \cite{wei2022chain} have hypothesized that generating intermediate steps expands an LM's ability to reason beyond a single forward pass \cite{nye2021show}; however, CFT suggests that we have not yet exhausted what can be done within one forward pass. Considering this optimal use of smaller models, CFT can be potentially used for ``distilling'' a complex multi-step workflow based on very large LMs---as seen in \cite{10.1145/3491102.3517582}---into one smaller LM.

We believe that our findings motivate research in fully automating the steps behind CFT. For the Decompose step, prior NLP works in decomposition \cite{dan-etal-2021-compositional-data, sakaguchi-etal-2021-proscript-partially, perez-etal-2020-unsupervised} could be expanded to this context. \citet{zhou2022least}, in particular, point to an interesting direction with ``least-to-most prompting.'' We note that the automation of the Decompose step is also warranted by chain of thought prompting, in which decomposition is also performed manually. For the Demonstrate step, automation would entail a few sub-steps: (i) exploring the lexical space (e.g., the space of possible preferences); (ii) generating paraphrases to increase natural language variation (e.g., our phrasings); and (iii) populating these phrasings with data from a domain of interest (i.e., $I^{full}$). In contrast, automating the Fine-Tune step is straightforward.

Finally, any models generated with CFT can be viewed as components themselves. For example, if a model is not able to handle larger sets of preferences or items (i.e., $|P|>1$ or $|I|>2$) in a decision template without losing performance, then one potential solution is to use an upstream agent to break a complex case into smaller ones (i.e., with $|P|=1$ and $|I|=2$) and combine their outputs. \citet{khot-etal-2022-hey} propose a framework than can be applied to this end. 

\section{Conclusion \& Future Work}
\label{sec:conclusion}

In this work, we proposed CFT as an improvement upon end-to-end learning. To enable research on this topic, we developed a new schema for generating recommendation datasets, which we instantiated in two domains. We showed that CFT indeed {\em consistently} outperforms end-to-end learning, as much as 32\% for local dining. Furthermore, we found evidence suggesting that more component tasks can be beneficial for CFT. Finally, instantiating chain of thought prompting in our dataset and CFT in sports understanding, we found CFT to be as good or better with LMs only 7.4\% of the size.

For future work, we plan to apply CFT to tasks with even more depth and breadth as in Figure \ref{fig:component_tasks}, as well as to conventional spatial navigation datasets (e.g., SCAN from \citet{lake2018generalization}). At the same time, we encourage others to test CFT on the large family of tasks that fit the inference types already covered in Figure \ref{fig:component_tasks}: those including facts, comparisons, criteria interpretations, and decisions, as seen in the recommendation task; or those including facts and assertions, as seen in sports understanding. We also plan to explore ways for fully automating the Decompose and Demonstrate steps. 

\section{Limitations}

This work focuses on testing if CFT outperforms end-to-end learning and chain of thought prompting in two very different domains. Despite the positive evidence, it remains to be seen: (i) if task decomposition can be fully automated, and (ii) if different decompositions---in the case of tasks that allow for multiple decompositions---yield similar results. Both are second-order research questions that can be pursued once compositionality has been confirmed to improve performance. Importantly, both questions have been left open in the initial chain of thought work as well. We hope that our results will add to theirs in attracting more attention to these questions in the future.

Another limitation of this work is that CFT is not applicable to several decomposition datasets that have been proposed. For example, a dataset focused on compositional generalization may include many different types of questions, each requiring different types of intermediate steps. CFT is not designed for intermediate steps that carry out very heterogeneous logic. Nonetheless, as shown in the recommendation tasks, CFT is still relevant for a substantial family of tasks with real-world applicability.

Lastly, this work is limited by its focus on the English language, and by the use of GPT-3 for its unique range of model sizes. For example, when we discuss that CFT on a 13B parameter model (Curie) is a much cheaper alternative to chain of thought prompting on a 175B parameter model (DaVinci), the finding is limited to this setting. It is important to replicate this work on other languages and models, which we plan to do as these become available.

\section*{Acknowledgements}

We would like to thank reviewers for their helpful feedback. This work was supported in part by gift funding from Adobe Research and by NSF grant IIS-2006851.

\bibliography{anthology,emnlp}
\bibliographystyle{acl_natbib}

\end{document}